\documentclass[final,5p,times,twocolumn]{elsarticle}

\usepackage{amssymb}

\usepackage{svg}
\usepackage{algorithm}
\usepackage{algpseudocode}
\usepackage{enumitem}
\usepackage{url}
\usepackage{algorithm}
\usepackage{algpseudocode}
\usepackage{enumitem}

\usepackage{textcomp}
\usepackage{stfloats}
\usepackage{hyperref}
\usepackage{multirow}
\usepackage{amsmath}

\usepackage{makecell}    

\newlist{mybulletlist}{itemize}{1}
\setlist[mybulletlist,1]{
  label=\textbullet,       
  left=0pt,               
  labelsep=10pt,          
  parsep=0pt,             
  itemsep=3pt,            
  before=\hspace{2em},    
  after=\hspace{2em}      
}

\journal{Neurocomputing}

\begin{document}

\begin{frontmatter}



\title{DRL-CLBA: A Clean Label Backdoor Attack for Speech Classification via DDPG Reinforcement Learning}


\author[1]{Yueming Huang}

\author[1]{Wenhan Yao}

\author[1]{Fen Xiao}

\author[2]{Xiarun Chen}

\author[2]{Weiping Wen\corref{cor1}}

\cortext[cor1]{Corresponding author}
\affiliation[1]{organization={Xiangtan University},
            addressline={Yuhu District Xiangda Road}, 
            city={Xiangtan},
            postcode={411100}, 
            state={Hunan},
            country={China}}
\affiliation[2]{organization={Peking University},
            addressline={No.5, Summer Palace Road, Haidian District, Beijing, China}, 
            city={Beijing},
            postcode={100871}, 
            state={Beijing},
            country={China}}        

\begin{abstract}
Deep learning models for speech classification are vulnerable to backdoor attacks, where malicious triggers cause misclassification at inference time. While sample-specific attacks can bypass many defenses, they often rely on poisoned label attack, making them detectable via manual data defense. In this paper, we propose DRL-CLBA, a novel clean label backdoor attack for speech classification that leverages Deep Deterministic Policy Gradient (DDPG) reinforcement learning. We also utilize deep audio steganography to embed sample-specific triggers into source audio, creating feature-space anchors. The proposed reinforcement learning framework effectively optimizes target samples toward trigger-bearing anchor points in the model’s deep latent space, enabling label-migration-free poisoning of target samples. Experimental results across three datasets and four different DNNs  demonstrate that DRL-CLBA achieves a high attack success rate , effectively bypassing some backdoor defenses. The attack demonstrates strong resistance against fine-tuning, pruning, and spectral signature defenses, exposing critical vulnerabilities in speech-controlled systems.
\end{abstract}


\end{frontmatter}



\section{Introduction}

Deep learning models have been extensively deployed in various voice-based human-computer interaction scenarios, such as the speech classification tasks: Keyword Spotting (KWS) \cite{berg2021keyword} and Speaker Verification (SV) \cite{desplanques2020ecapa}. To improve generalization performance in real-world environments, practitioners frequently incorporate synthetic noise into training datasets for data augmentation. However, existing research indicates that such noise can be exploited by adversaries as a potential threat vector, specifically for backdoor attacks \cite{li2022backdoor}.

Backdoor attacks are generally categorized into poisoned label and clean label  attacks. In the poisoned label paradigm, an adversary injects a specific trigger into a subset of benign training data and modifies their corresponding labels to a target class. This process establishes a strong correlation between the trigger and the target class during the training stage. At the inference stage, the model functions as a backdoored model, which misclassifies poisoned samples into the target class, while still providing correct predictions for benign inputs.  Gu et al. \cite{gu2019badnets} pioneered this field with BadNets, the first backdoor attack against deep image classification models, which utilized pixel-block patterns as triggers. Subsequent research has introduced diverse trigger patterns \cite{liu2020reflection,hammoud2022check,chen2021badnl}. However, as these triggers are typically independent of the underlying sample content, they remain vulnerable to backdoor defenses designed for trigger filtering \cite{wang2019neural}. Most conventional methods require the simultaneous manipulation of both samples and labels. This allows potential victims to thwart attacks by auditing training datasets to detect label-content inconsistencies or identifying low-level image editing artifacts.

Clean label backdoor attacks represent a more pragmatic and stealthy threat, as they maliciously tamper with training samples without altering their ground-truth labels. This class of attack typically begins by introducing perturbations into samples of a target class to generate "poisoned samples." These perturbations are designed to shift the poisoned samples' representations within the feature space of a deep classification model, aligning them more closely with samples containing a trigger (referred to as backdoor samples). Consequently, the model is coerced into learning a semantic similarity between the backdoor samples and the poisoned samples, leading it to misclassify triggered inputs into the target category. Shen et al. proposed CSSBA \cite{shen2023cssba}, which utilizes image steganography models for trigger implantation and employs the Projected Gradient Descent (PGD) algorithm to iteratively optimize poisoned samples through feature collision. Similarly, Cai et al. \cite{cai2025clean} attempted to facilitate attacks by weakening the inherent robust features of data samples within a surrogate model, subsequently transferring the attack to the target model. However, these methodologies require access to the full gradient information of the deep classification model, which is often inaccessible to attackers in real-world scenarios. Furthermore, these attacks frequently employ fixed-pattern triggers during the inference phase, rendering them susceptible to detection by defensive mechanisms.

To uncover more latent security threats in  speech recognition models, this paper proposes a novel attack methodology: Deep Deterministic action Gradient-based clean label  Backdoor Attack (\textbf{DRL-CLBA}). This approach operates without requiring full gradient access to the target classification model, demonstrating both superior stealth and attack efficacy. In the attack phase, we employ advanced audio deep steganography to embed triggers into source class audio, generating backdoor images that serve as anchors for feature collisions. The use of steganographic techniques ensures that our triggers are sample-specific rather than following a static pattern. Subsequently, an actor-critic network is utilized to optimize the generation of poisoned samples. Specifically, the Actor network produces optimized differentials for poisoned samples based on the current feature collision distance and deep feature representations. Meanwhile, the Critic network employs a reward function to maximize long-term feature collision gains, incorporating label-indistinguishability and semantic consistency rewards to achieve a clean label  attack.

The optimization strategy for these networks is governed by the Deep Deterministic action Gradient (DDPG) algorithm, which is a type of reinforcement learning (RL) algorithm. By ensuring the feature vectors of the optimized poisoned samples are nearly identical to those of the backdoor samples, the model is compelled to learn the backdoor association. Notably, the adversary is not required to modify the labels of any samples.

We summarize our contributions as follows:
\begin{itemize}
    \item We propose a clean label , sample-specific backdoor attack that achieves a high attack success rate while effectively bypassing conventional backdoor defense mechanisms.

    \item  We leverage deep steganography to generate backdoor speechs and integrate them with feature collision techniques to produce poisoned samples, ensuring high imperceptibility.

    \item The effectiveness of the proposed attack is rigorously validated across multiple datasets and speech classification architectures.

\end{itemize}

\section{Background}

\subsection{Backdoor Attacks}

The 360 Vulnerability Research Institute released the \textit{Large Model Security Vulnerability Report} in 2024, highlighting the risk of backdoor injection in large language models. The report documents an incident in which Hugging Chat Assistants was subjected to a backdoored model replacement, triggering information exfiltration when users entered email-related inputs \cite{360-2024-llm-security}. A joint micro-poisoning study conducted by institutions including the UK AI Safety Institute demonstrates that as few as 250 malicious documents are sufficient to successfully implant a backdoor into a large model, causing it to generate nonsensical outputs when specific trigger words are provided \cite{souly2025poisoning}. These reports indicate that studying how backdoor attacks pose threats in real-world scenarios is of significant importance.

As illustrated in Figure \ref{fig:后门攻击步骤_cut}, a complete backdoor attack implementation typically comprises a series of algorithmic components, including the (1) predefined number of backdoors, (2) trigger function design, (3) poisoning strategies, (4) model training protocols, and (5) attack effectiveness evaluation. The attacker first determines the number of backdoors to be implanted, which directly influences the subsequent algorithm design. In the trigger function design, implanting a one target backdoor is common in image classification tasks. For example, triggers can take the form of white pixels \cite{gu2019badnets}, rose \cite{wang2025versatile}, reflection patterns \cite{chen2017targeted}, eyeglasses, or color transformations \cite{jiang2024rethinking}. However, single-target backdoor attacks can be mitigated by backdoor defense methods such as STRIP \cite{gao2019strip}, model pruning \cite{liu2018fine}, and Neural Cleanse \cite{wang2019neural}; therefore, attackers tend to implant multiple backdoors into the model. The effectiveness of multi-backdoor attacks largely depends on the design of trigger functions that can flexibly control multiple target behaviors. Xue et al. \cite{xue2020one} first introduced one-to-N and N-to-one trigger mappings, demonstrating that trigger functions can encode complex activation logic across multiple labels. Subsequent works explored more advanced trigger designs. Doan et al. \cite{doan2022marksman} and Salem et al. \cite{salem2022dynamic} modeled trigger functions as class-conditional generators, enabling adaptive and data-dependent trigger patterns for different targets. Hou et al. \cite{hou2024m} further proposed an M-to-N paradigm, allowing multiple triggers to activate multiple target classes. Meanwhile, Li et al. \cite{li2024multi} enhanced attack diversity by employing heterogeneous triggers in parallel or sequential poisoning processes. Overall, these studies show that increasingly flexible and learnable trigger function designs are key to achieving scalable and stealthy multi-backdoor attacks. Typically, once the trigger function for a backdoor attack is designed, the attacker carries out a poisoning strategy by applying it to clean samples together with the trigger to generate poisoned samples, and then modifies the clean labels to target labels, forming a poisoned set to contaminate the training dataset. However, there also exist alternative poisoning strategies that do not modify the clean labels to target labels, which poses a significant challenge for the model to learn the association between the trigger and the target label.

\begin{figure}[t]
    \centering
    \includegraphics[width=1\linewidth]{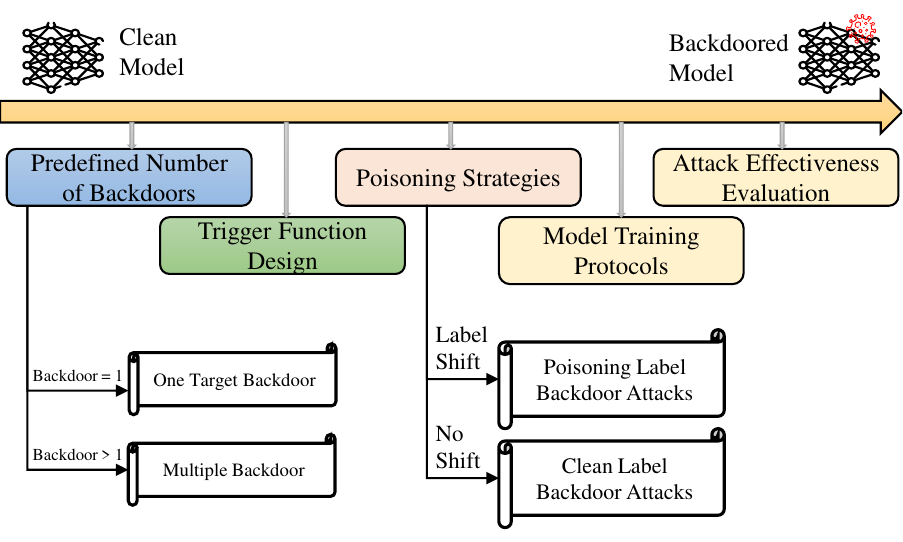}
    \caption{General backdoor attack algorithm workflow}
    \label{fig:cut}
\end{figure}

Clean-label backdoor attacks avoid modifying labels, making them more stealthy but also harder to optimize, especially in multi-backdoor settings. Existing works mainly extend single-backdoor strategies to support multiple triggers through different optimization mechanisms. First, adversarial optimization-based methods enable multiple trigger learning without label supervision. Gong et al. \cite{gong2025megatron} and Wang et al. \cite{wang2024clean} generate optimized triggers via surrogate models, while Xing et al. \cite{xinyuan2024clean} and Yin et al. \cite{yin2025ffcba} leverage PGD or generative models to learn diverse trigger patterns, which can be naturally extended to multi-trigger scenarios. Second, feature manipulation-based methods enhance trigger effectiveness by weakening clean features or amplifying trigger signals. Zhu et al. \cite{zhu2025towards} and You et al. \cite{you2025ultimate} utilize attribute editing to construct multiple semantic triggers, while Xie et al. \cite{xie2025clean} and Yuan et al. \cite{yuan2025stealthy} suppress original features in spatial or frequency domains, enabling multiple triggers to coexist with reduced interference. Finally, sample selection-based methods focus on poisoning boundary or sensitive samples to improve multi-trigger reliability. Wu et al. \cite{wu2025clear} and Tang et al. \cite{tang2025cats} identify vulnerable samples near decision boundaries and embed adaptive triggers, while Zhang et al. \cite{zhang2025clean} further enhances multi-trigger separability through feature alignment strategies. Overall, these studies suggest that combining clean-label constraints with diverse trigger generation, feature manipulation, and sample selection strategies provides a feasible pathway toward scalable multi-backdoor attacks.

\subsection{Speech Classification Models}

This research focuses on speech classifiers. Speech classifiers predict labels based on the speech waveforms or spectrograms. In recent years, deep neural networks (DNNs) have achieved optimal efficacy in the domains \cite{berg2021keyword,choi2019temporal,huang2023limi,bartoli2025end,xi2025ntc}, hence raising substantial security issues. The classifier $C_{\theta}$ is consistently tuned using the cross-entropy loss as outlined below:

\begin{align}
        L_{(x,y) \in D} = \mathop{\arg\max}\limits_{\theta} p(y \mid C_{\theta}(x)) 
\end{align}

The $(x,y)$ represents the model inputs and corresponding true labels. Training on the dataset $D$ enables the classifier to discern the correlations between a particular input feature and the potential attack labels, hence posing a risk of data poisoning backdoor attacks.

\section{Methodology}

\subsection{Threat Model}

Commonly, backdoor attacks occur when an adversary conceals a backdoored model behind an API or distributes a compromised model to end-users, thereby facilitating malicious exploitation. 

We consider a clean label threat model in which the attacker is allowed to manipulate the training dataset and influence the training process, but has no access to modifying the architecture or internal parameters of the target model. In particular, all injected training samples must be correctly labeled according to their semantic content, making the attack difficult to detect through label inspection. Under this constraint, the attacker aims to implant a backdoor into the model by carefully crafting malicious yet benign-looking samples during training.

The primary goal of the attacker is to maximize the attack success rate (ASR), defined as the probability that inputs containing the trigger are misclassified into the target class at inference time, while maintaining a high benign accuracy (BA). Here, benign accuracy refers to the classification accuracy on clean, non-poisoned test images. Preserving BA is essential to ensure that the backdoored model exhibits normal performance on benign data and thus avoids suspicion. In addition, the invisibility of malicious images constitutes a critical requirement in the clean label setting. The perturbations or triggers embedded in poisoned samples are designed to be imperceptible to human observers and to remain semantically consistent with the original class, further enhancing the stealthiness of the attack and reducing the likelihood of detection by manual inspection or automated defense mechanisms.

\subsection{The Pipeline of Proposed Attack.}

Figure \ref{fig:1-DRL-CLBAmethod} illustrates the injection pipeline of our proposed attack, which comprises four primary stages: (1) Initialization Step, (2) DRL-CLBA Training, (3) Poisoned Dataset Generation, and (4) Backdoor Implantation. In the first step, we initialize the various neural network architectures required for reinforcement learning. Simultaneously, an audio steganography model is utilized to generate backdoor samples and their corresponding feature vectors, which serve as anchors for the subsequent optimization. In the second step, we employ a DDPG-based reinforcement learning framework to maximize a series of reward functions, specifically focusing on feature collision. This trains the actor network to generate poisoned samples that are precisely feature collisions with the target anchors. In the third step, the optimized poisoned samples are injected into the target training set following the standard protocol for poisoning-based backdoor attacks. Critically, we maintain the clean label  nature of the attack by preserving the original labels of all speech samples. In the last step, the target model undergoes training on the poisoned training dataset, resulting in the convergence of a backdoored model.

\subsection{The Principle of Feature Collision}
\label{chapter:Feature Collision}
\begin{figure}[t]
    \centering
    \includegraphics[scale=0.6]{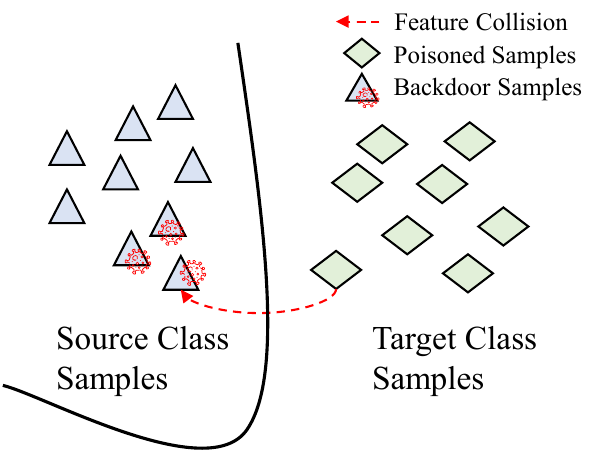}
    \caption{Feature collisions}
    \label{fig:2-featurecoll}
\end{figure}

Figure \ref{fig:2-featurecoll} illustrates the mechanism of the clean label backdoor attack via feature collision. Let the feature extraction component of the target DNN model be denoted by the function $h(\cdot)$. Let the source class sample set be $X_{src} = \{x_i^s\}_{i=1}^{K_{src}}$ and the target class sample set be $X_{tar} = \{x_i^t\}_{i=1}^{K_{tar}}$.  First, the adversary utilizes a trigger function $\xi$ to construct backdoor samples—specifically, samples embedded with a trigger $\delta_{trg}$, denoted as $\xi(x_i^s, \delta_{trg})$. The resulting feature vectors extracted by the DNN, $H_{trg}^{src} = \{h_i^{src} = h(\xi(x_i^s, \delta_{trg}) ) \}$, serve as anchors. Subsequently, the adversary employs a feature collision algorithm to optimize a poisoned sample $\hat{x}_i^t \leftarrow x_i^t$. The optimization objective is to minimize the distance between the feature representation of the poisoned sample, $h(\hat{x}_i^t)$, and a specific anchor in $H_{trg}^{src}$. Consequently, the DNN is compelled to learn the following mapping: $h(\xi(x_i^s, \delta_{trg})) \approx h(\hat{x}_i^t)$. Thus, the model consequently misidentifies the predicted labels of source-class samples embedded with triggers as highly similar to the optimized target-class labels in the deep latent space. Based on this property, an attacker can exploit the optimized target-class samples to successfully carry out the attack.

In the data poisoning stage, the adversary then injects a significant volume of these poisoned samples $\hat{x}_i^t$ into a clean training dataset to execute the poisoning-based backdoor attack. During the inference phase, the adversary applies the trigger to source class samples; the DNN subsequently misclassifies these samples into the target category, signifying a successful backdoor activation.
\subsection{ Initialization Step and Deep Steganography}

The initial stage consists of two primary steps. In the first step, we utilize an audio deep steganography model, referred to as the backdoor generator, to produce anchor vectors. In the second step, we initialize the components of the Deep Deterministic action Gradient (DDPG) framework: the Actor network $\mu(s | \theta^\mu)$ and the Critic (Value) network $Q(s, a | \theta^Q)$. Furthermore, to facilitate the estimation of long-term feature collision gains and ensure training stability, we initialize two corresponding target networks, $\mu'$ and $Q'$, with identical weights to their primary counterparts.

The backdoor generator is designed as an encoder-decoder architecture based on deep steganography \cite{ge2025cross}, comprising three functional modules: the preparation network, the hiding network, and the reveal network. The preparation network encodes the trigger audio into a lower-dimensional representation. The hiding network is an 1D convolutional encoder that accepts both the preparation network's output and the source class audio sample $x_s$ as inputs, producing the backdoor sample $x_s^{trg}$. The reveal network decodes the latent information from the backdoor sample to reconstruct the original trigger audio, outputting the predicted trigger $\delta'_{trg}$. By leveraging an encoder-decoder structure with high-density residual connections, the generator ensures that the steganographic triggers are sample-specific—varying for each backdoor sample rather than following a fixed pattern. The training objective for the backdoor generator is defined by the following loss function:
\begin{equation}
    L_g(x_s, x_s^{trg}, \delta_{trg}, \delta'_{trg}) = \|x_s - x_s^{trg}\| + \|\delta_{trg} - \delta'_{trg}\|
\end{equation}
The first term aims to reduce the differences between source audios and backdoor audios, thereby ensuring the imperceptibility of the backdoor attack. The second term is to minimizing the reconstruction error of the trigger audios.

\begin{figure*}[ht]
    \centering
    \includegraphics[width=1\linewidth]{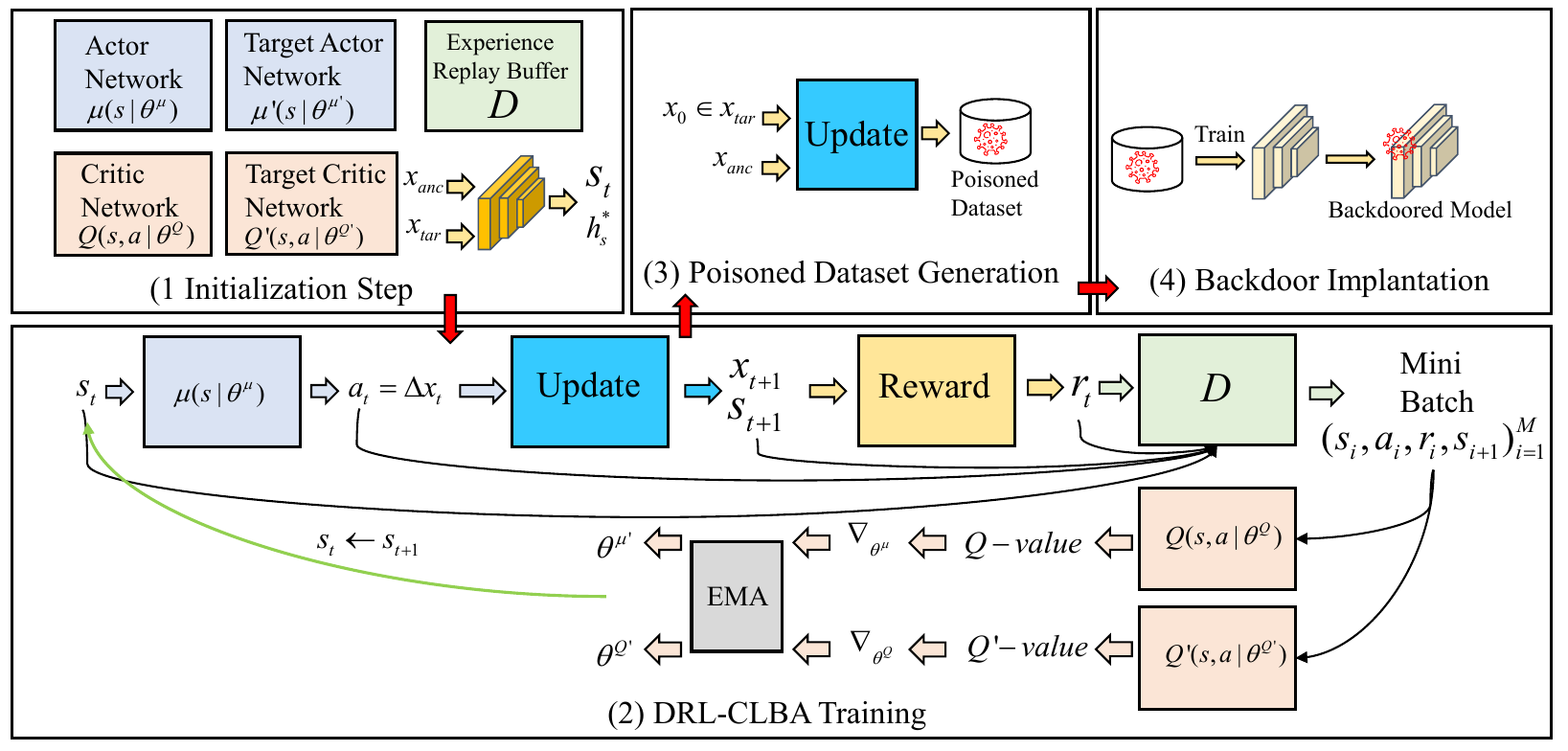}
    \caption{The DRL-CLBA attack pipeline}
    \label{fig:1-DRL-CLBAmethod}
\end{figure*}

\subsection{DRL-CLBA Training}

The DRL-CLBA training process is illustrated in Figure \ref{fig:1-DRL-CLBAmethod}(2). To address the limitations of gradient based feature collision optimization, we reformulate clean label backdoor attacks as a sequential decision making process. We guide the generation of poisoned samples through a Markov Decision Process (MDP) defined by the quadruple $(S, A, P, R)$. In this formulation, the gradual perturbation of a target class sample, which aimed at aligning its deep feature representation with a trigger embedded source class samples, is solved using the Deep Deterministic action Gradient (DDPG) algorithm. Next, we formalize the state, reward, action network, and value network in the MDP.

\noindent\textbf{State, Action, Update.} At each decision step $t$, the classification model acts as an agent observing a state $s_t \in S$, which captures the current alignment between the poisoned sample and the source-class anchor. To ensure stability and avoid high-dimensional raw data issues, the state is constructed using deep feature representations and constraint-related statistics:
\begin{equation}
    s_t = \big[ h(x_t),\; h_s^*,\; \| h(x_t) - h_s^* \|_2,\; \| x_t - x_{tar} \|_2 \big]
\end{equation}
where $h(x_t)$ represents the fixed feature extractor's output for the $t$ step intermediate poisoned sample $x_t$, and $h_s^*$ denotes one of anchor features of the triggered source class sample. The $x_{tar}$ denotes a target class sample. This design explicitly accounts for the dual objectives of feature collision and perturbation boundedness. The agent interacts with the environment to obtain gradient information that characterizes how the current sample should be updated in order to optimize the feature collision distance. Based on this information, the agent applies a continuous perturbation increment $\Delta x_t$ to iteratively update the poisoned sample. Specifically, the sample is updated as
\begin{equation}
x_{t+1} = \Pi_{\epsilon}(x_t + \Delta x_t),
\end{equation}
where $\Pi_{\epsilon}$ denotes a projection operator that enforces the constraint $|x_{t+1} - x_{tar}|_p \leq \epsilon$. This projection ensures that the generated poisoned sample does not deviate excessively from the initial target point. Moreover, this iterative update mechanism enables a multi-step optimization process that incorporates historical interaction information, rather than relying on a single-step update strategy, thereby improving the stability and flexibility of the attack optimization procedure.

\noindent\textbf{Reward.} The optimization is governed by a multi-objective reward function $r_t$ designed to balance feature collision, imperceptibility, and semantic consistency. It is defined as:
\begin{multline}
r_t = - \lambda_1\| h(x_{t+1}) - h_s^* \|_2 - \lambda_2 \| x_{t+1} - x_{tar} \|_2 \\ + \lambda_3 \cdot \mathbb{I}[f(x_{t+1}) = y_{tar}]
\end{multline}
By penalizing the distance to the anchor features and the original target sample while rewarding correct target class predictions $y_{tar}$, the reward function enables the agent to learn sophisticated attack strategies that are often difficult to optimize using traditional gradient descent.

\noindent\textbf{Action and Value Network.} The attack strategy employs a Deep Deterministic Policy Gradient (DDPG) framework. The action network $\mu(s | \theta^\mu)$ maps the state $s$ to a continuous perturbation $\Delta x_t$, perceiving the joint distribution of feature and input spaces to drive samples toward the target domain. To evaluate these actions, the value network $Q(s, a | \theta^Q)$ estimates the expected cumulative reward, balancing immediate collision gains with future potential.To ensure training stability and prevent divergence, the architecture incorporates target networks $\mu'$ and $Q'$, synchronized through soft updates. Structurally, the value network is a Multi-Layer Perceptron (MLP) that fuses state and action vectors into a scalar quality assessment, directly guiding the gradient updates for the action network.

\noindent\textbf{The DDPG Training Procedure.} The training procedure involves a cycle of interaction, sampling, and network refinement. First, the agent interacts with the environment to store transitions $(s_t, a_t, r_t, s_{t+1})$ in a replay buffer $\mathcal{D}$. Subsequently, a mini-batch of $N$ samples is drawn to update the networks. The value network is optimized by minimizing the Mean Squared Error (MSE) loss between the predicted Q-value and the target value $y_i$:
\begin{equation}
    L(\theta^Q) = \frac{1}{N} \sum_i \left( y_i - Q(s_i, a_i | \theta^Q) \right)^2
\end{equation}
where the target $y_i$ is computed as:
\begin{equation}
    y_i = r_i + \gamma Q'(s_{i+1}, \mu'(s_{i+1} | \theta^{\mu'}) | \theta^{Q'})
\end{equation}
This loss function ensures that the critic provides an accurate assessment of the "quality" of a perturbation strategy.Simultaneously, the policy network is updated using the Deterministic Policy Gradient Theorem, aiming to maximize the expected value as evaluated by the critic. The gradient for the policy parameters $\theta^\mu$ is approximated as:
\begin{equation}
    \resizebox{0.4\textwidth}{!}{$
        \nabla_{\theta^\mu} J \approx \mathbb{E}_{s_t \sim \mathcal{D}} \left[ \nabla_a Q(s, a | \theta^Q) \big|_{s=s_t, a=\mu(s_t)} \nabla_{\theta^\mu} \mu(s | \theta^\mu) \big|_{s=s_t} \right]
    $}
\end{equation}
Finally, the target networks are updated via an exponential moving average (EMA) with a decay rate $\tau \ll 1$, ensuring the convergence and stability of the entire optimization process across iterations. In our method, a multilayer perceptron is used to implement the action and value networks.

\subsection{Poisoned Dataset Generation}

Once the DRL training phase is complete, the optimized policy $\mu^*$ is frozen and deployed as a generator to produce poisoned samples from the target class dataset $\mathcal{X}_{tar}$. Unlike traditional optimization-based attacks, this generation process is an iterative inference procedure that does not require access to the model's gradients. Given an initial target class sample $x_0 \in \mathcal{X}_{tar}$, an anchor $h_s^*$ is first selected based on the nearest neighbor principle. This forms the initial state $s_0 = \big[ h(x_0),\; h_s^*,\; \| h(x_0) - h_s^* \|_2,\; 0 \big]$. The poisoned sample is then evolved through a sequence of deterministic actions:
\begin{equation}
    x_{t+1} = \Pi_{\epsilon} (x_t + \mu^*(s_t | \theta^\mu)), \quad t=0, 1, \dots, T
\end{equation}
This process continues until a maximum number of steps $T$ is reached or the feature distance $d_t$ falls below a predefined threshold. Notably, the state input for the policy network depends solely on the output of the model’s final feature layer rather than its internal weight gradients. This significantly relaxes the requirement for the attacker's surrogate model to match the target model’s architecture or parameters, potentially enabling clean label attacks in scenarios where only model outputs are accessible.

The final output of the inference process is defined as the poisoned sample $x_{poi} = x_T$. Due to the projection operator used during iteration, the sample maintains visual consistency with the original target label $y_{tar}$ in the pixel space, satisfying $\|x_{poi} - x_0\|_p \leq \epsilon$.To construct the complete poisoned subset $\mathcal{D}_{poi}$, the frozen policy $\mu^*$ is applied to a randomly selected portion of the target class dataset $\mathcal{X}_{tar}$ governed by the poisoning rate $\rho_{poi}$. This results in a subset $\mathcal{D}_{poi} = \{ (x_{poi}^i, y_{tar}^i) \}_{i=1}^{N_{\rho}}$, where $N_{\rho}$ denotes the number of poisoned samples. Crucially, the original labels $y_{tar}$ are preserved throughout this process, ensuring the "clean label" nature of the attack, which bypasses manual data inspection.

\subsection{Backdoor Implantation}

In the final stage, the attacker merges the poisoned subset $\mathcal{D}_{poi}$ with the original benign training dataset $\mathcal{D}_{benign}$ to form the final training dataset $\mathcal{D}_{train} = \mathcal{D}_{benign} \cup \mathcal{D}_{poi}$. The victim model is then trained on $\mathcal{D}_{train}$ using standard supervised learning procedures. No modifications to the model's hyperparameters or loss functions are required, as the model updates its parameters $\theta$ by minimizing the standard cross-entropy loss:
\begin{equation}
\min_{\theta} \; \sum_{(x_i, y_i) \in \mathcal{D}_{train}}
\mathcal{L}_{ce}(M(x_i; \theta), y_i)
\end{equation}
Although the training process itself remains conventional, the underlying data distribution has been strategically manipulated. Because $x_{poi}$ is forced into the feature distribution region of the source category (i.e., $h(x_{poi}) \approx h_s^*$), the model is compelled to associate the source category's feature space with the target label $y_{tar}$ to achieve minimal training error. This "feature-level misleading" creates a latent backdoor vulnerability; while the model performs normally on clean data, it will misclassify any source-class sample containing the trigger as the target class during deployment.

\section{Experiments and Results}

\subsection{Models and Datasets. } 

Our experiments were conducted on three different speech classification tasks: keyword spotting (KWS), speaker verification (SV), and speech emotion recognition (SER). For the KWS task, we use three different KWS datasets, including SCD \cite{warden2018speech}, AudioMNIST \cite{becker2024audiomnist}, and LibriKWS-20 \cite{xi2024tdt}. The first two datasets contain 35 and 10 classes of commands, respectively. LibriKWS-20 is a subset of the evaluation split of the LibriSpeech dataset and consists of 20 classes. For the SV task, we use the AISHELL3-50 and VoxCeleb1-50 \cite{nagrani2017voxceleb} datasets. VoxCeleb1-50 is a SV dataset derived from the large-scale VoxCeleb1 corpus, which was collected from YouTube videos. It contains 18354 utterances from the top 50 most talkative speakers selected from the VoxCeleb1-H subset. AISHELL3-50 is a SV dataset extracted from the AISHELL-3 Mandarin corpus. It comprises 18081 utterances from the 50 speakers with the most recordings. For the SER task, we split the English and Chinese datasets of the ESD corpus \cite{zhou2022emotional} into two separate sub-datasets. They share the same number of speakers and utterances. Each dataset contains 10 speakers with 5 emotional states.

Our attacks target four DNN models: ERes2Net \cite{chen2023enhanced}, KWS-ViT \cite{berg2021keyword}, EAT-S \cite{gazneli2022end}, and CAM++ \cite{wang2023cam++}. All four models are suitable for speech classification tasks and respectively cover traditional convolutional architectures, advanced self-attention mechanisms, as well as their combinations.

\subsection{Baseline Selection.} 

Following the methodology described in chapter 
\ref{chapter:Feature Collision}, we adapt existing non-clean label poisoning based single backdoor attacks into clean label attack variants, utilizing the Projected Gradient Descent (PGD) algorithm to optimize the target samples. The adapted methodologies include the Ultra method \cite{koffas2022can} and the OneSpec method \cite{zhai2021backdoor}. Furthermore, this study extends mature clean label attack frameworks to speech classification models, incorporating the following three advanced approaches: Clean Label Backdoor Attack (CBA) \cite{turner2018clean}, Clean Label Sample-Specific Backdoor Attack (CSSBA) \cite{shen2023cssba}, Targeted Universal Adversarial Perturbation Backdoor Attack (TUAPBA) \cite{cai2025clean}.

\begin{table*}[ht]
\centering
\scriptsize
\caption{Attack performance on the KWS task}
\label{table:5.6.1-kws}
\begin{tabular}{cccccccccc}
\hline
\multicolumn{1}{c|}{\multirow{2}{*}{Dataset}}    & \multicolumn{1}{c|}{\multirow{2}{*}{Model}}       & \multicolumn{1}{c|}{\makecell[{{c}}]{\rule{0pt}{1.2em} Metrics \\ (BA$\uparrow$ \\ ASR$\uparrow$)}} & \multicolumn{1}{c|}{\multirow{2}{*}{Acc}} & \multicolumn{5}{c|}{Baselines}   & Proposed                 \\ \cline{5-10} 
\multicolumn{1}{c|}{}                        & \multicolumn{1}{c|}{}                          & \multicolumn{1}{c|}{}                    & \multicolumn{1}{c|}{}                     & Ultra                  & OneSpec              &  CBA              & CSSBA         & \multicolumn{1}{c|}{TUAPBA}  & DRL-CLBA              \\ \hline
\multicolumn{1}{c|}{\multirow{8}{*}{SCD}} & \multicolumn{1}{c|}{\multirow{2}{*}{ERes2Net}} & \multicolumn{1}{c|}{BA}                  & \multicolumn{1}{c|}{93.42}                & 94.12                & 92.12              & 93.55               & 93.07               & \multicolumn{1}{c|}{93.15} & 93.76              \\
\multicolumn{1}{c|}{}                        & \multicolumn{1}{c|}{}                          & \multicolumn{1}{c|}{ASR}                 & \multicolumn{1}{c|}{-}                    & 34.67                & 42.15                & 64.15               & 76.34               & \multicolumn{1}{c|}{85.22} & 88.12              \\ \cline{2-10} 
\multicolumn{1}{c|}{}                        & \multicolumn{1}{c|}{\multirow{2}{*}{KWS-ViT}}  & \multicolumn{1}{c|}{BA}                  & \multicolumn{1}{c|}{94.08}                & 92.12              & 93.54                & 91.86               & 92.17               & \multicolumn{1}{c|}{92.01} & 93.95              \\
\multicolumn{1}{c|}{}                        & \multicolumn{1}{c|}{}                          & \multicolumn{1}{c|}{ASR}                 & \multicolumn{1}{c|}{-}                    & 31.27                & 45.78                & 58.76               & 72.14                & \multicolumn{1}{c|}{78.69} & 85.78            \\ \cline{2-10} 
\multicolumn{1}{c|}{}                        & \multicolumn{1}{c|}{\multirow{2}{*}{EAT-S}}    & \multicolumn{1}{c|}{BA}                  & \multicolumn{1}{c|}{94.29}                &95.11               & 93.17               & 93.20                & 94.28                & \multicolumn{1}{c|}{95.01} & 94.89             \\
\multicolumn{1}{c|}{}                        & \multicolumn{1}{c|}{}                          & \multicolumn{1}{c|}{ASR}                 & \multicolumn{1}{c|}{-}                    & 33.42               & 40.15               & 45.17                & 66.45             & \multicolumn{1}{c|}{72.73} & 91.02           \\ \cline{2-10} 
\multicolumn{1}{c|}{}                        & \multicolumn{1}{c|}{\multirow{2}{*}{CAM++}}    & \multicolumn{1}{c|}{BA}                  & \multicolumn{1}{c|}{93.54}                & 89.17               & 88.64               & 90.56               & 92.24              & \multicolumn{1}{c|}{92.74} & 93.60             \\
\multicolumn{1}{c|}{}                        & \multicolumn{1}{c|}{}                          & \multicolumn{1}{c|}{ASR}                 & \multicolumn{1}{c|}{-}                    & 36.17               & 46.78               & 50.16               & 73.12              & \multicolumn{1}{c|}{76.12} & 87.45              \\

\hline



\multicolumn{1}{c|}{\multirow{8}{*}{AudioMNIST}} & \multicolumn{1}{c|}{\multirow{2}{*}{ERes2Net}} & \multicolumn{1}{c|}{BA}                  & \multicolumn{1}{c|}{95.02}                & 94.89                & 95.21                & 93.99               & 94.58               & \multicolumn{1}{c|}{94.72} & 96.18             \\
\multicolumn{1}{c|}{}                        & \multicolumn{1}{c|}{}                          & \multicolumn{1}{c|}{ASR}                 & \multicolumn{1}{c|}{-}                    & 39.50                & 51.87               & 63.46                & 78.15               & \multicolumn{1}{c|}{82.18} & 91.10 \\ \cline{2-10} 
\multicolumn{1}{c|}{}                        & \multicolumn{1}{c|}{\multirow{2}{*}{KWS-ViT}}  & \multicolumn{1}{c|}{BA}                  & \multicolumn{1}{c|}{97.72}                & 96.87               & 95.48               & 96.28                & 97.67                & \multicolumn{1}{c|}{97.55} & 96.98             \\
\multicolumn{1}{c|}{}                        & \multicolumn{1}{c|}{}                          & \multicolumn{1}{c|}{ASR}                 & \multicolumn{1}{c|}{-}                    & 42.33                & 52.17                & 60.89              & 80.68                & \multicolumn{1}{c|}{81.54} & 89.48            \\ \cline{2-10} 
\multicolumn{1}{c|}{}                        & \multicolumn{1}{c|}{\multirow{2}{*}{EAT-S}}    & \multicolumn{1}{c|}{BA}                  & \multicolumn{1}{c|}{96.18}                & 95.34              & 94.27               & 96.46              & 97.01              & \multicolumn{1}{c|}{96.29} & 96.88              \\
\multicolumn{1}{c|}{}                        & \multicolumn{1}{c|}{}                          & \multicolumn{1}{c|}{ASR}                 & \multicolumn{1}{c|}{-}                    & 46.89              & 55.87               & 72.18                & 85.15               & \multicolumn{1}{c|}{88.77} & 93.15              \\ \cline{2-10} 
\multicolumn{1}{c|}{}                        & \multicolumn{1}{c|}{\multirow{2}{*}{CAM++}}    & \multicolumn{1}{c|}{BA}                  & \multicolumn{1}{c|}{95.67}                & 95.25                & 96.34               & 95.88                & 96.22               & \multicolumn{1}{c|}{96.74} & 95.89             \\
\multicolumn{1}{c|}{}                        & \multicolumn{1}{c|}{}                          & \multicolumn{1}{c|}{ASR}                 & \multicolumn{1}{c|}{-}                    & 52.14              & 46.78               & 63.47                & 75.85                & \multicolumn{1}{c|}{82.44} & 86.78               \\

\hline


\multicolumn{1}{c|}{\multirow{8}{*}{LibriKWS-20}} & \multicolumn{1}{c|}{\multirow{2}{*}{ERes2Net}} & \multicolumn{1}{c|}{BA}                  & \multicolumn{1}{c|}{93.17}                & 93.20                & 92.21                & 94.55                & 89.07              & \multicolumn{1}{c|}{88.85} & 92.47              \\
\multicolumn{1}{c|}{}                        & \multicolumn{1}{c|}{}                          & \multicolumn{1}{c|}{ASR}                 & \multicolumn{1}{c|}{-}                    & 17.28                & 49.15               & 55.87                & 78.15               & \multicolumn{1}{c|}{80.01} & 86.71 \\ \cline{2-10} 
\multicolumn{1}{c|}{}                        & \multicolumn{1}{c|}{\multirow{2}{*}{KWS-ViT}}  & \multicolumn{1}{c|}{BA}                  & \multicolumn{1}{c|}{92.25}                & 91.56               & 90.88               & 90.16               & 92.08               & \multicolumn{1}{c|}{90.31} & 92.34              \\
\multicolumn{1}{c|}{}                        & \multicolumn{1}{c|}{}                          & \multicolumn{1}{c|}{ASR}                 & \multicolumn{1}{c|}{-}                    & 28.79               & 35.46               & 46.58                & 63.8               & \multicolumn{1}{c|}{78.28} & 87.64             \\ \cline{2-10} 
\multicolumn{1}{c|}{}                        & \multicolumn{1}{c|}{\multirow{2}{*}{EAT-S}}    & \multicolumn{1}{c|}{BA}                  & \multicolumn{1}{c|}{95.08}                & 95.89               & 96.12              & 96.74              & 95.78               & \multicolumn{1}{c|}{96.14} & 95.10           \\
\multicolumn{1}{c|}{}                        & \multicolumn{1}{c|}{}                          & \multicolumn{1}{c|}{ASR}                 & \multicolumn{1}{c|}{-}                    & 21.07              & 40.15               & 45.77                & 58.15               & \multicolumn{1}{c|}{70.57} & 85.21               \\ \cline{2-10} 
\multicolumn{1}{c|}{}                        & \multicolumn{1}{c|}{\multirow{2}{*}{CAM++}}    & \multicolumn{1}{c|}{BA}                  & \multicolumn{1}{c|}{96.65}                & 97.15                & 96.75             & 95.71                & 96.65              & \multicolumn{1}{c|}{96.17} & 97.54             \\
\multicolumn{1}{c|}{}                        & \multicolumn{1}{c|}{}                          & \multicolumn{1}{c|}{ASR}                 & \multicolumn{1}{c|}{-}                    & 31.87             & 39.17                & 55.78                & 47.12                & \multicolumn{1}{c|}{86.57} & 91.07                \\ 
\hline
\end{tabular}
\end{table*}


\begin{table*}[ht]
\centering
\scriptsize
\caption{Attack performance on the SV task}
\label{table:5.6.1-sv}
\begin{tabular}{cccccccccc}
\hline
\multicolumn{1}{c|}{\multirow{2}{*}{Dataset}}    & \multicolumn{1}{c|}{\multirow{2}{*}{Model}}       & \multicolumn{1}{c|}{\makecell[{{c}}]{\rule{0pt}{1.2em} Metrics \\ (BA$\uparrow$ \\ ASR$\uparrow$)}} & \multicolumn{1}{c|}{\multirow{2}{*}{Acc}} & \multicolumn{5}{c|}{Baselines}   & Proposed                \\ \cline{5-10} 
\multicolumn{1}{c|}{}                        & \multicolumn{1}{c|}{}                          & \multicolumn{1}{c|}{}                    & \multicolumn{1}{c|}{}                     & Ultra                  & OneSpec              &  CBA              & CSSBA         & \multicolumn{1}{c|}{TUAPBA}  & DRL-CLBA              \\ \hline
\multicolumn{1}{c|}{\multirow{8}{*}{AISHELL3-50}} & \multicolumn{1}{c|}{\multirow{2}{*}{ERes2Net}} & \multicolumn{1}{c|}{BA}                  & \multicolumn{1}{c|}{95.17}                & 94.84               & 95.84              & 96.55               & 95.75             & \multicolumn{1}{c|}{95.15} & 96.76              \\
\multicolumn{1}{c|}{}                        & \multicolumn{1}{c|}{}                          & \multicolumn{1}{c|}{ASR}                 & \multicolumn{1}{c|}{-}                    & 40.67                & 42.15                & 55.47               & 75.34               & \multicolumn{1}{c|}{88.17} & 92.14             \\ \cline{2-10} 
\multicolumn{1}{c|}{}                        & \multicolumn{1}{c|}{\multirow{2}{*}{KWS-ViT}}  & \multicolumn{1}{c|}{BA}                  & \multicolumn{1}{c|}{96.08}                & 97.34           & 97.54                & 96.55               & 96.17               & \multicolumn{1}{c|}{97.01} & 97.55              \\
\multicolumn{1}{c|}{}                        & \multicolumn{1}{c|}{}                          & \multicolumn{1}{c|}{ASR}                 & \multicolumn{1}{c|}{-}                    & 28.12                & 37.78                & 60.72               & 68.41              & \multicolumn{1}{c|}{85.60} & 90.78            \\ \cline{2-10} 
\multicolumn{1}{c|}{}                        & \multicolumn{1}{c|}{\multirow{2}{*}{EAT-S}}    & \multicolumn{1}{c|}{BA}                  & \multicolumn{1}{c|}{94.19}                &95.55               & 93.46               & 93.82                & 94.67               & \multicolumn{1}{c|}{93.01} & 94.25            \\
\multicolumn{1}{c|}{}                        & \multicolumn{1}{c|}{}                          & \multicolumn{1}{c|}{ASR}                 & \multicolumn{1}{c|}{-}                    & 43.51             & 50.15               & 67.52                & 70.45             & \multicolumn{1}{c|}{80.79} & 85.73          \\ \cline{2-10} 
\multicolumn{1}{c|}{}                        & \multicolumn{1}{c|}{\multirow{2}{*}{CAM++}}    & \multicolumn{1}{c|}{BA}                  & \multicolumn{1}{c|}{93.54}                & 89.17               & 90.17               & 91.51               & 93.24              & \multicolumn{1}{c|}{92.54} & 93.01            \\
\multicolumn{1}{c|}{}                        & \multicolumn{1}{c|}{}                          & \multicolumn{1}{c|}{ASR}                 & \multicolumn{1}{c|}{-}                    & 49.17               & 38.88               & 51.11               & 65.55              & \multicolumn{1}{c|}{75.12} & 86.44              \\

\hline



\multicolumn{1}{c|}{\multirow{8}{*}{VoxCeleb1-50}} & \multicolumn{1}{c|}{\multirow{2}{*}{ERes2Net}} & \multicolumn{1}{c|}{BA}                  & \multicolumn{1}{c|}{92.02}                & 92.69                & 91.55                & 91.89               & 92.98               & \multicolumn{1}{c|}{91.80} & 92.86             \\
\multicolumn{1}{c|}{}                        & \multicolumn{1}{c|}{}                          & \multicolumn{1}{c|}{ASR}                 & \multicolumn{1}{c|}{-}                    & 42.50                & 60.87               & 55.46                & 75.15               & \multicolumn{1}{c|}{83.18} & 89.52 \\ \cline{2-10} 
\multicolumn{1}{c|}{}                        & \multicolumn{1}{c|}{\multirow{2}{*}{KWS-ViT}}  & \multicolumn{1}{c|}{BA}                  & \multicolumn{1}{c|}{94.72}                & 95.46               & 93.38               & 94.75               & 94.87               & \multicolumn{1}{c|}{94.12} & 95.70             \\
\multicolumn{1}{c|}{}                        & \multicolumn{1}{c|}{}                          & \multicolumn{1}{c|}{ASR}                 & \multicolumn{1}{c|}{-}                    & 55.30                & 58.67               & 63.71              & 75.45                & \multicolumn{1}{c|}{75.54} & 88.12            \\ \cline{2-10} 
\multicolumn{1}{c|}{}                        & \multicolumn{1}{c|}{\multirow{2}{*}{EAT-S}}    & \multicolumn{1}{c|}{BA}                  & \multicolumn{1}{c|}{95.18}                & 95.57              & 94.64               & 95.38              & 94.01              & \multicolumn{1}{c|}{95.28} & 94.58              \\
\multicolumn{1}{c|}{}                        & \multicolumn{1}{c|}{}                          & \multicolumn{1}{c|}{ASR}                 & \multicolumn{1}{c|}{-}                    & 51.89              & 67.87               & 62.18                & 72.15               & \multicolumn{1}{c|}{88.50} & 91.15              \\ \cline{2-10} 
\multicolumn{1}{c|}{}                        & \multicolumn{1}{c|}{\multirow{2}{*}{CAM++}}    & \multicolumn{1}{c|}{BA}                  & \multicolumn{1}{c|}{92.67}                & 91.25                & 90.76               & 93.83                & 92.14               & \multicolumn{1}{c|}{92.50} & 92.71            \\
\multicolumn{1}{c|}{}                        & \multicolumn{1}{c|}{}                          & \multicolumn{1}{c|}{ASR}                 & \multicolumn{1}{c|}{-}                    & 33.11              & 25.68               & 54.48                & 65.88                & \multicolumn{1}{c|}{81.40} & 90.70              \\ 
\hline
\end{tabular}
\end{table*}


\begin{table*}[h]
\centering
\scriptsize
\caption{Attack performance on the SER task}
\label{table:5.6.1-SER}
\begin{tabular}{cccccccccc}
\hline

\multicolumn{1}{c|}{\multirow{2}{*}{Dataset}}    & \multicolumn{1}{c|}{\multirow{2}{*}{Model}}       & \multicolumn{1}{c|}{\makecell[{{c}}]{\rule{0pt}{1.2em} Metrics \\ (BA$\uparrow$ \\ ASR$\uparrow$)}} & \multicolumn{1}{c|}{\multirow{2}{*}{Acc}} & \multicolumn{5}{c|}{Baselines}   & Proposed                 \\ \cline{5-10} 
\multicolumn{1}{c|}{}                        & \multicolumn{1}{c|}{}                          & \multicolumn{1}{c|}{}                    & \multicolumn{1}{c|}{}                     & Ultra                  & OneSpec              &  CBA              & CSSBA         & \multicolumn{1}{c|}{TUAPBA}  & DRL-CLBA              \\ \hline
\multicolumn{1}{c|}{\multirow{8}{*}{ESD-CN}} & \multicolumn{1}{c|}{\multirow{2}{*}{ERes2Net}} & \multicolumn{1}{c|}{BA}                  & \multicolumn{1}{c|}{91.64}                & 90.24                & 91.17             & 90.45               & 90.54               & \multicolumn{1}{c|}{92.15} & 90.17              \\
\multicolumn{1}{c|}{}                        & \multicolumn{1}{c|}{}                          & \multicolumn{1}{c|}{ASR}                 & \multicolumn{1}{c|}{-}                    & 25.45              & 33.45                & 45.15               & 50.34               & \multicolumn{1}{c|}{71.22} & 85.51             \\ \cline{2-10} 
\multicolumn{1}{c|}{}                        & \multicolumn{1}{c|}{\multirow{2}{*}{KWS-ViT}}  & \multicolumn{1}{c|}{BA}                  & \multicolumn{1}{c|}{93.48}                & 94.12              & 92.54                & 92.45              & 93.17              & \multicolumn{1}{c|}{93.01} & 93.57             \\
\multicolumn{1}{c|}{}                        & \multicolumn{1}{c|}{}                          & \multicolumn{1}{c|}{ASR}                 & \multicolumn{1}{c|}{-}                    & 17.64                & 28.64                & 35.45               & 45.65               & \multicolumn{1}{c|}{55.69} & 76.63            \\ \cline{2-10} 
\multicolumn{1}{c|}{}                        & \multicolumn{1}{c|}{\multirow{2}{*}{EAT-S}}    & \multicolumn{1}{c|}{BA}                  & \multicolumn{1}{c|}{94.55}                &91.56               & 92.17               & 93.25                & 92.18              & \multicolumn{1}{c|}{93.01} & 92.89             \\
\multicolumn{1}{c|}{}                        & \multicolumn{1}{c|}{}                          & \multicolumn{1}{c|}{ASR}                 & \multicolumn{1}{c|}{-}                    & 18.92               & 27.52               & 39.87                & 42.64             & \multicolumn{1}{c|}{56.73} & 75.48           \\ \cline{2-10} 
\multicolumn{1}{c|}{}                        & \multicolumn{1}{c|}{\multirow{2}{*}{CAM++}}    & \multicolumn{1}{c|}{BA}                  & \multicolumn{1}{c|}{95.54}                & 95.16               & 96.54               & 94.45              & 95.77              & \multicolumn{1}{c|}{94.64} & 95.25             \\
\multicolumn{1}{c|}{}                        & \multicolumn{1}{c|}{}                          & \multicolumn{1}{c|}{ASR}                 & \multicolumn{1}{c|}{-}                    & 12.55              & 34.57              & 45.16               & 50.12              & \multicolumn{1}{c|}{62.12} & 72.17              \\

\hline



\multicolumn{1}{c|}{\multirow{8}{*}{ESD-EN}} & \multicolumn{1}{c|}{\multirow{2}{*}{ERes2Net}} & \multicolumn{1}{c|}{BA}                  & \multicolumn{1}{c|}{90.42}                & 92.12                & 91.05               & 92.13              & 90.12              & \multicolumn{1}{c|}{90.25} & 90.68             \\
\multicolumn{1}{c|}{}                        & \multicolumn{1}{c|}{}                          & \multicolumn{1}{c|}{ASR}                 & \multicolumn{1}{c|}{-}                    & 30.51                & 28.87               & 45.47               & 63.85               & \multicolumn{1}{c|}{72.14} & 83.51 \\ \cline{2-10} 
\multicolumn{1}{c|}{}                        & \multicolumn{1}{c|}{\multirow{2}{*}{KWS-ViT}}  & \multicolumn{1}{c|}{BA}                  & \multicolumn{1}{c|}{92.45}                & 92.89               & 93.37               & 91.26                & 92.69                & \multicolumn{1}{c|}{92.45} & 92.68             \\
\multicolumn{1}{c|}{}                        & \multicolumn{1}{c|}{}                          & \multicolumn{1}{c|}{ASR}                 & \multicolumn{1}{c|}{-}                    & 19.33                & 28.17                & 47.89              & 55.68                & \multicolumn{1}{c|}{71.54} & 85.56           \\ \cline{2-10} 
\multicolumn{1}{c|}{}                        & \multicolumn{1}{c|}{\multirow{2}{*}{EAT-S}}    & \multicolumn{1}{c|}{BA}                  & \multicolumn{1}{c|}{93.18}                & 94.37              & 94.26               & 95.49              & 93.51              & \multicolumn{1}{c|}{94.99} & 92.56              \\
\multicolumn{1}{c|}{}                        & \multicolumn{1}{c|}{}                          & \multicolumn{1}{c|}{ASR}                 & \multicolumn{1}{c|}{-}                    & 14.41              & 28.35       
& 38.18                & 42.15               & \multicolumn{1}{c|}{72.54} & 81.15              \\ \cline{2-10} 
\multicolumn{1}{c|}{}                        & \multicolumn{1}{c|}{\multirow{2}{*}{CAM++}}    & \multicolumn{1}{c|}{BA}                  & \multicolumn{1}{c|}{91.1}                & 90.25                & 91.54               & 90.98                & 91.52               & \multicolumn{1}{c|}{91.25} & 91.07             \\
\multicolumn{1}{c|}{}                        & \multicolumn{1}{c|}{}                          & \multicolumn{1}{c|}{ASR}                 & \multicolumn{1}{c|}{-}                    & 20.14              & 25.78               & 38.48                & 59.85                & \multicolumn{1}{c|}{69.44} & 82.78               \\ 
\hline

\end{tabular}
\end{table*}

\subsection{Evaluation Results and Comparison}

\noindent\textbf{Attack Setup.} On three datasets, we set the poisoning rate $\rho_{poi}=0.8\%$. For our attack, we use the single frequency audio from the OneSpec as the trigger audio and perform feature collision for 3000 epochs.
For the backdoor generator, we use the adam optimizer to fintune 80 epochs on all datasets. The initial learning rate is 0.0002, and the learning rate is divided by 0.97 for every 10 epochs.

\noindent\textbf{Attack Performance. }  \textbf{(1) KWS task.} Table \ref{table:5.6.1-kws} presents the experimental results of clean-label backdoor attacks on three KWS datasets across four different deep neural network models. The proposed DRL-CLBA method is compared with five baseline attack methods. The evaluation metrics include ASR and BA. On the KWS task, the DRL-CLBA method achieves average ASR values of 88.09\%, 89.76\%, and 90.12\% on the SCD, AudioMNIST, and LibriKWS-20 datasets, respectively. Compared with single backdoor settings where ASR can easily exceed 98\%, these results verify that clean-label attacks are significantly more challenging. However, across all three datasets and models, the ASR of the DRL-CLBA method consistently exceeds the maximum ASR achieved by the baseline methods. The differences between the ASR of DRL-CLBA and the average ASR of the five baseline methods on the three datasets are 31.62\%, 31.11\%, and 25.01\%, respectively. This demonstrates that, compared with the standard PGD algorithm, the reinforcement-learning-driven poisoned samples proposed in this work are more effectively learned by the model as incorrect deep features. From the experimental results, it can also be observed that different triggers used as anchor samples lead to significantly different outcomes. For example, when using the Ultra method as a trigger, the ASR values on the three datasets are 33.88\%, 37.85\%, and 45.21\%, respectively. When using the OneSpec method as a trigger, the ASR values are 43.71\%, 47.74\%, and 51.67\%, respectively. This indicates that these two triggers struggle to form effective anchor points in the deep feature space to guide poisoned samples. In contrast, the TUAPBA, CSSBA, and CBA methods achieve average ASR values of 80.26\%, 71.26\%, and 56.85\% across all datasets, respectively. This suggests that TUAPBA, which jointly applies sample feature weakening and deep feature collision, is effective, but still does not surpass the proposed DRL-CLBA method. These experimental results validate that the proposed steganographic model can form effective anchor features in the deep feature space, and also demonstrate the effectiveness of reinforcement-learning-driven poisoned sample optimization. \textbf{(2) SV task.} Table \ref{table:5.6.1-sv} presents the experimental results of clean-label backdoor attacks on two SV datasets (AISHELL3-50 and VoxCeleb1-50) across four deep neural network models. The proposed DRL-CLBA method is compared with five baseline methods, and the evaluation metrics include ASR and BA. On the SV task, the DRL-CLBA method achieves average ASR values of 88.77\% and 87.45\% on the AISHELL3-50 and VoxCeleb1-50 datasets, respectively. The mean performance of the proposed method exceeds the maximum ASR achieved by all baseline methods, indicating that the proposed approach remains effective on other tasks. Similar to the results on the speech command recognition task, the Ultra and OneSpec methods still yield relatively low ASR values, with averages not exceeding 58\%. In contrast, the TUAPBA, CSSBA, and CBA methods achieve average ASR values of 58.83\%, 71.04\%, and 82.28\%, respectively. This suggests that these three methods—using adversarial perturbations, steganographic triggers, and frequency-band noise respectively—are more effective than directly manipulating the frequency-domain space. The experimental results validate that the proposed DRL-CLBA method still achieves a satisfactory attack performance on the SV task and consistently outperforms all baseline methods. \textbf{(3) SER task.} The experimental results of clean-label backdoor attacks on SER tasks are presented in Table \ref{table:5.6.1-SER}. This experiment is conducted on two mainstream speech emotion recognition datasets (ESD-CN and ESD-EN), and compares four deep neural network models to evaluate the performance of the proposed DRL-CLBA method against baseline methods in terms of clean-label attack effectiveness. The evaluation metrics include ASR and BA. On the speech emotion recognition task, the average ASR of the proposed DRL-CLBA method on the ESD-CN and ESD-EN datasets is 77.44\% and 79.18\%, respectively. This attack success rate is lower than that observed in speech command recognition and speaker recognition tasks, which may be attributed to the more dispersed nature of the emotional speech feature space. Among the baseline methods, Ultra and OneSpec exhibit similar behavior, both achieving very low ASR values, with average ASR of 19.86\% and 29.41\%, respectively. This indicates that, for speech emotion recognition tasks, triggers in clean-label attacks need to be robust against emotional feature representations. In contrast, methods such as CBA, CSSBA, and TUAPBA, which incorporate adversarial perturbations and feature weakening strategies, achieve average ASR values of 41.95\%, 51.28\%, and 66.42\%, respectively. This further validates that the trigger injection strategy has a significant impact on clean-label backdoor attacks. Overall, the experimental results demonstrate that the proposed DRL-CLBA method can still achieve satisfactory attack success rates on speech emotion recognition tasks and consistently outperforms existing baseline methods.

\begin{figure}[h]
    \centering
    \includegraphics[scale=0.70]{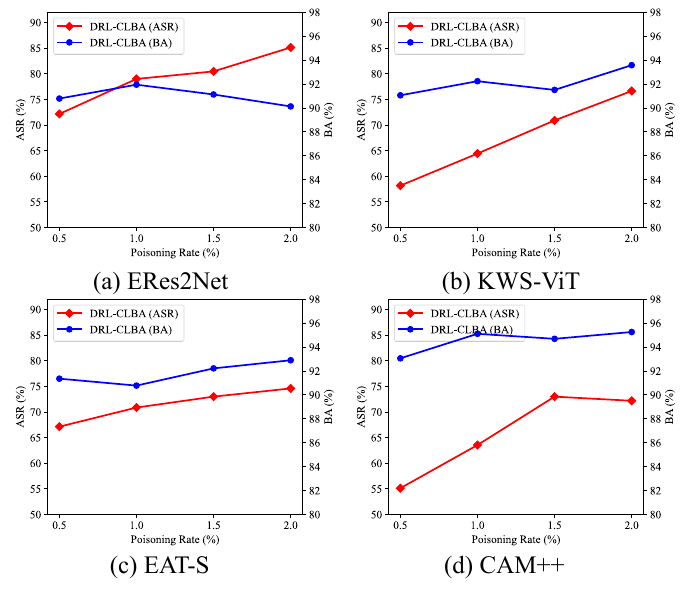}
    \caption{Impact of different poisoning rates on DRL-CLBA}
    \label{fig:diff-poirate}
\end{figure}

\noindent\textbf{Impact of the Poisoning Rate.} The effect of the poisoning rate on DRL-CLBA was evaluated on the SCD dataset using four architectures. As the poisoning rate increased from 0.5\% to 2.0\%, all models maintained high BA (typically above 80\%), indicating minimal impact on normal classification and strong stealthiness. Meanwhile, ASR increased monotonically with the poisoning rate, surpassing 65\% at rates below 1.0\% and gradually saturating as the rate approached 2.0\. DRL-CLBA exhibited stable performance across both CNN- and Transformer-based models, with lightweight CNNs achieving higher ASR and Transformer models showing stronger benign robustness. Overall, DRL-CLBA achieves an effective trade-off between attack effectiveness and stealth while demonstrating strong cross-model generalization.

\begin{figure}[ht]
    \centering
    \includegraphics[scale=0.70]{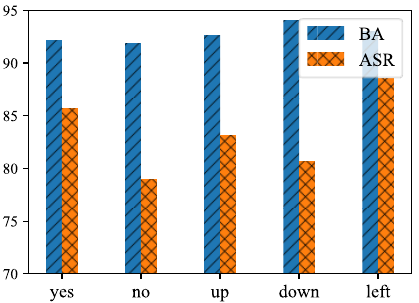}
    \caption{Impact of different target labels on DRL-CLBA}
    \label{fig:diff-targetlabel}
\end{figure}

\noindent\textbf{Impact of Target Labels.} To assess generality, five target labels were randomly selected on SCD. ASR varies notably across targets, peaking at 89.6\% for Left and dropping to 78.9\% for No, reflecting differences in class feature separability and alignment with model decision boundaries. These results indicate that target label selection plays a critical role in balancing attack effectiveness and stealth.

\begin{table}[h]
\centering
\caption{Impact of reward terms on DRL-CLBA}
\label{table-5.6.2-消融-奖励函数目标}
\small
\setlength{\tabcolsep}{4pt}
\begin{tabular}{c|c|cccc}
\hline
\makecell{Reward\\Terms} & Metrics$\uparrow$ & ERes2Net & KWS-ViT & EAT-S & CAM++ \\ \hline
\multirow{2}{*}{\textbf{all}} & ASR & 88.12 & 85.78 & 91.02 & 87.45 \\
 & BA & 93.76 & 93.95 & 94.89 & 93.60 \\ \hline
\multirow{2}{*}{\textbf{w/o pc}} & ASR & 89.14 & 86.57 & 91.04 & 88.14 \\
 & BA & 90.18 & 89.47 & 91.47 & 90.85 \\ \hline
\multirow{2}{*}{\textbf{w/o sp}} & ASR & 74.78 & 78.87 & 85.41 & 82.14 \\
 & BA & 92.15 & 93.15 & 92.48 & 92.54 \\ \hline
\end{tabular}
\end{table}

\noindent\textbf{Impact of Reward Functions.} 
We investigate the impact of removing components other than feature collision. The experiments include: (1) using all components (\textbf{all}), (2) removing the perturbation constraint $\| x_{t+1} - x_{tar} \|_2$ (\textbf{w/o pc}),  and (3) eliminating semantic preservation $\beta \cdot \mathbb{I}[f(x_{t+1}) = y_{tar}]$ \textbf{(w/o sp)}. Removing the perturbation constraint leads to a significant decline in BA, notably dropping to 89.47\% in KWS-ViT, indicating that excessive visual distortion impairs model generalization and exposes the attack to manual audit. Conversely, eliminating the semantic preservation term causes the ASR to plummet, such as the drop from 88.12\% to 74.78\% in ERes2Net, proving that feature collision alone is insufficient if the sample deviates from the target class's semantic distribution. These results confirm that the synergy of all three components is indispensable.

\noindent\textbf{Impact of Attack Robustness.}  Clean-label attacks typically assume prior knowledge of the victim model architecture, i.e., a white-box setting. Although the proposed method is applicable to black-box scenarios, we further evaluate its robustness across different surrogate models. Experimental design: (1) Consistent attack–inference models: ERes2Net is used in both the attack and inference stages. (2) Inconsistent attack–inference models: The policy is trained using ERes2Net or EAT-S as the surrogate model to generate poisoned samples during the attack stage, while the target model at inference is ResNet-50. (3) Feature-level impact: The anchor point $h^{*}_{s}$
is changed from the “penultimate-layer feature” to an “intermediate-layer feature”. This ablation study aims to verify whether the perturbation strategy learned by the DRL-CLBA policy network has general “feature-level intervention” capability, rather than merely overfitting to specific network weights. All experiments are conducted on the SCD dataset.

The ablation experiments of DRL-CLBA demonstrate that: under white-box conditions, the attack success rate reaches 88\%-91\%, while maintaining a high success rate of 77\%-84\% during black-box cross-model transfer, indicating strong generalization capability across different model architectures. Furthermore, the attack effectiveness significantly degrades as the feature hierarchy deepens (decreasing from 72\% to 23\%), suggesting that the learned strategy primarily relies on high-level semantic features for effective perturbation rather than shallow local responses.

\noindent\textbf{Impact of Sequence Decision Length.} Similar to the PGD method, which iteratively updates adversarial examples, DRL-CLBA is also an iterative process. Table 5.6 illustrates the Attack Success Rate (ASR) of the DRL-CLBA clean-label attack on the SCD dataset using the ERes2Net model under different sequential decision lengths (i.e., the number of poisoning iterations $T$).The results indicate that as $T$ increases, the ASR for all three methods rises significantly, proving that iterative optimization positively impacts attack performance. However, DRL-CLBA consistently achieves the best results across almost all iteration counts. At $T=1$, its ASR is only 0.89\%, slightly lower than CSSBA (0.97\%) and TUAPBA (1.05\%), suggesting that DRL-CLBA’s sequential modeling advantage is not yet apparent in a single-step decision phase. As $T$ increases to 10 and 20, DRL-CLBA rapidly overtakes other methods (reaching 2.47\% and 6.78\% respectively), demonstrating stronger initial convergence.Notably, in the high-iteration phase ($T \ge 60$), DRL-CLBA maintains a substantial lead, reaching 72.84\% at $T=80$ and 83.45\% at $T=100$, far exceeding CSSBA (78.15\%) and TUAPBA (76.12\%). This suggests that the Reinforcement Learning framework used by DRL-CLBA effectively captures long-term dependencies, optimizing the generation path of poisoned samples through sequential decisions to achieve more efficient adversarial perturbation accumulation. In contrast, while CSSBA and TUAPBA possess iterative capabilities, they lack global strategic planning in complex decision spaces, leading to slower convergence and limited final performance.Remarkably, DRL-CLBA achieves a 46.78\% ASR at $T=40$, nearly matching or exceeding the performance of other methods at $T=100$. This validates the superiority of modeling the attack as an MDP: by introducing state representation and reward mechanisms, Reinforcement Learning can explore better action sequences and avoid local optima, resulting in a more robust and efficient attack path than traditional gradient methods.In conclusion, this ablation study demonstrates the critical impact of sequential decision length on ASR. DRL-CLBA significantly outperforms single-step strategies when $T > 1$ and maintains a leading edge in long sequences, strongly supporting the core hypothesis of "modeling clean-label attacks as a Markov Decision Process." The results show that the RL framework offers better adaptability and generalization when handling complex, non-linear attack optimization problems, providing theoretical and practical guidance for designing more efficient and stealthy adversarial attacks.

\begin{table}[t]
\centering
\caption{Impact of Sequence Decision Length on ASR}
\resizebox{\columnwidth}{!}{%
\begin{tabular}{c|ccccccc}
\hline
Methods & \multicolumn{7}{c}{Iteration of Target Poisoned Sample} \\ \hline
 & 1 & 10 & 20 & 40 & 60 & 80 & 100 \\
CSSBA    & 0.97 & 2.74 & 5.79 & 38.48 & 60.78 & 68.72 & 78.15 \\
TUAPBA   & 1.05 & 3.85 & 10.84 & 56.47 & 63.78 & 67.18 & 76.12 \\
DRL-CLBA & 0.89 & 2.47 & 6.78 & 46.78 & 62.78 & 72.84 & 83.45 \\
\hline
\end{tabular}%
}
\label{table-5.6.2}
\end{table}

\subsection{Resistance to Backdoor Defenses.}

Currently, many backdoor defense methods have been proposed to mitigate backdoor threats in image classification tasks \cite{guo2023scale,xiang2023umd,jebreel2023defending}. However, most of them cannot be directly applied to speech classification tasks, as they are specifically designed for the image domain. Therefore, in this section, we evaluate the proposed DRL-CLBA attack under four classic and representative cross-domain defense methods, including fine-tuning \cite{liu2017neural}, model pruning \cite{liu2018fine}, and spectral signatures \cite{tran2018spectral}. For simplicity, experiments in this section are conducted on the AudioMNIST dataset using the ERes2Net and EAT-S models.

\begin{figure}[ht]
    \centering
    \includegraphics[width=1\linewidth]{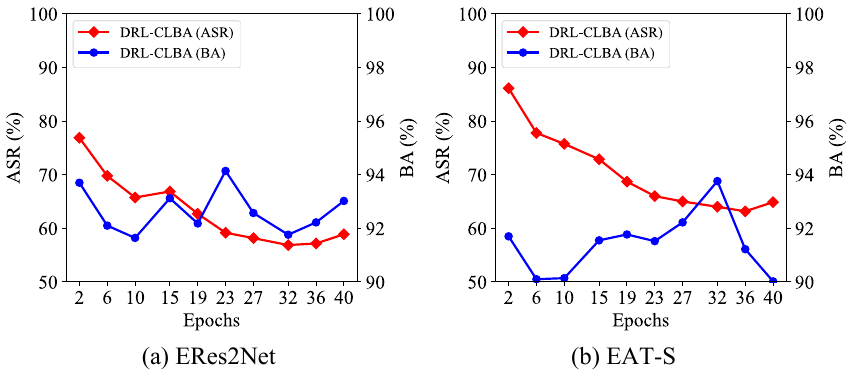}
    \caption{Resistance of DRL-CLBA to fine-tuning defenses.}
    \label{fig:4-defense-finetuning}
\end{figure}

\noindent\textbf{Resistance to Fine-tuning.} As illustrated in Figure \ref{fig:4-defense-finetuning}, as the number of fine-tuning epochs increases, the attack success rate (ASR) of the proposed DRL-CLBA method on both ERes2Net and EAT-S exhibits a decreasing trend, indicating that fine-tuning exerts a certain suppressive effect on the attack performance. Nevertheless, it is noteworthy that even after 40 fine-tuning epochs, DRL-CLBA still achieves an ASR of approximately 60\% on ERes2Net, while maintaining an ASR above 65\% on EAT-S. Meanwhile, the classification accuracy on clean samples remains consistently above 92\% throughout the entire fine-tuning process, suggesting that the attack does not significantly impair the model’s normal classification capability. These results demonstrate that DRL-CLBA possesses strong resilience against fine-tuning-based defenses, effectively preserving the embedded backdoor behavior during parameter updates. This robustness highlights the method’s stability in terms of both trigger effectiveness and attack persistence. In particular, although the ASR on EAT-S shows minor fluctuations during fine-tuning, its overall performance remains stable, further validating the effectiveness of DRL-CLBA in resisting common defense strategies.

\begin{figure}
    \centering
    \includegraphics[width=1\linewidth]{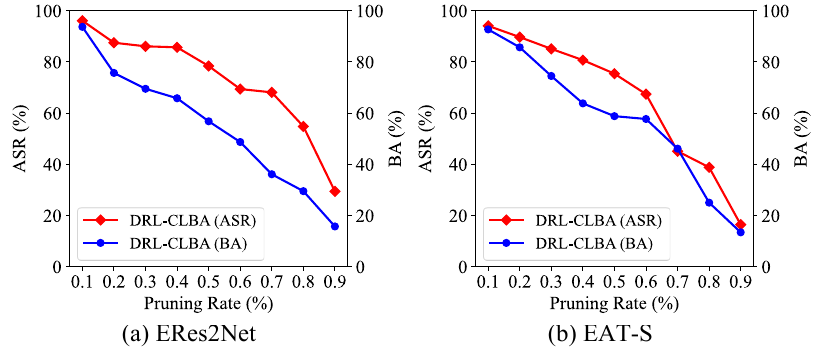}
    \caption{Resistance of DRL-CLBA to model pruning defenses.}
    \label{fig:4-defense-pruning}
\end{figure}

\noindent\textbf{Resistance to Model Pruning.} As shown in Figure \ref{fig:4-defense-pruning}, as the pruning rate increases, both the attack success rate (ASR) and the benign accuracy (BA) of the DRL-CLBA method exhibit a significant downward trend. Specifically, when the pruning rate is raised from 0.1\% to 0.9\%, the ASR and BA of both models (ERes2Net and EAT-S) decrease consistently, with highly synchronized magnitudes of degradation. Notably, in the high pruning rate regime ($> 0.6\%$ ), the decay patterns of ASR and BA become nearly identical, indicating that while pruning suppresses backdoor behaviors, it also severely impairs the model’s normal classification capability. This observation suggests that the DRL-CLBA method demonstrates strong resistance to pruning-based defense strategies: although pruning can weaken the backdoor effect, it does so at the cost of substantial overall performance degradation. Consequently, these results confirm that DRL-CLBA maintains high robustness against model fine-tuning type defense mechanisms.

\begin{figure}
    \centering
    \includegraphics[width=1\linewidth]{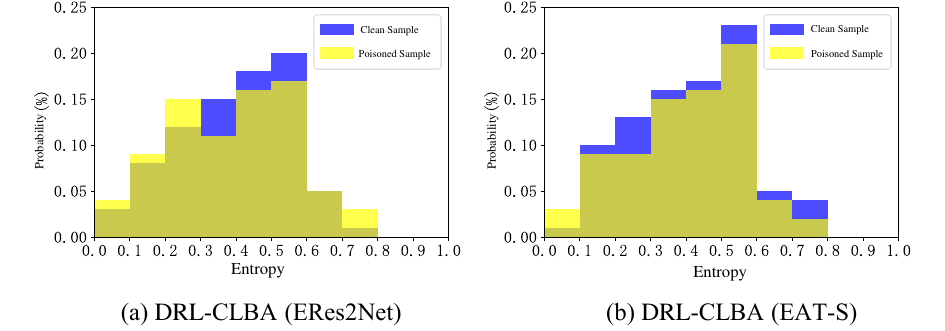}
    \caption{Resistance of DRL-CLBA to STRIP defenses.}
    \label{fig:4-defense-STRIP}
\end{figure}

\noindent\textbf{Resistance to STRIP.} As shown in Figure \ref{fig:4-defense-STRIP}, the experiment evaluates the resistance of the DRL-CLBA method against the STRIP defense mechanism by visualizing the entropy distribution of clean and poisoned samples at the model output. Comparative Analysis of Entropy Distribution On both the ERes2Net (Figure a) and EAT-S (Figure b) models, the entropy distributions of clean samples (blue) and poisoned samples (yellow) exhibit a high degree of overlap. Particularly in the high-entropy intervals, the distribution trends of the two types of samples are nearly indistinguishable. This indicates that the backdoor trigger constructed by DRL-CLBA does not introduce significant abnormal behavior or detectable statistical bias during the model's prediction process. Consequently, it effectively circumvents the STRIP detection mechanism, which relies on entropy differences for sample classification.  Furthermore, although poisoned samples show a slight shift in certain low-entropy regions, the overall distribution remains highly consistent with that of clean samples, demonstrating the strong stealthiness of the attack. The experimental results verify that while maintaining a high attack success rate, DRL-CLBA possesses robust resistance to STRIP detection, showcasing excellent stealthiness and robustness.

\subsection{t-SNE Analysis}

\begin{figure}[h]
    \centering
    \includegraphics[width=1\linewidth]{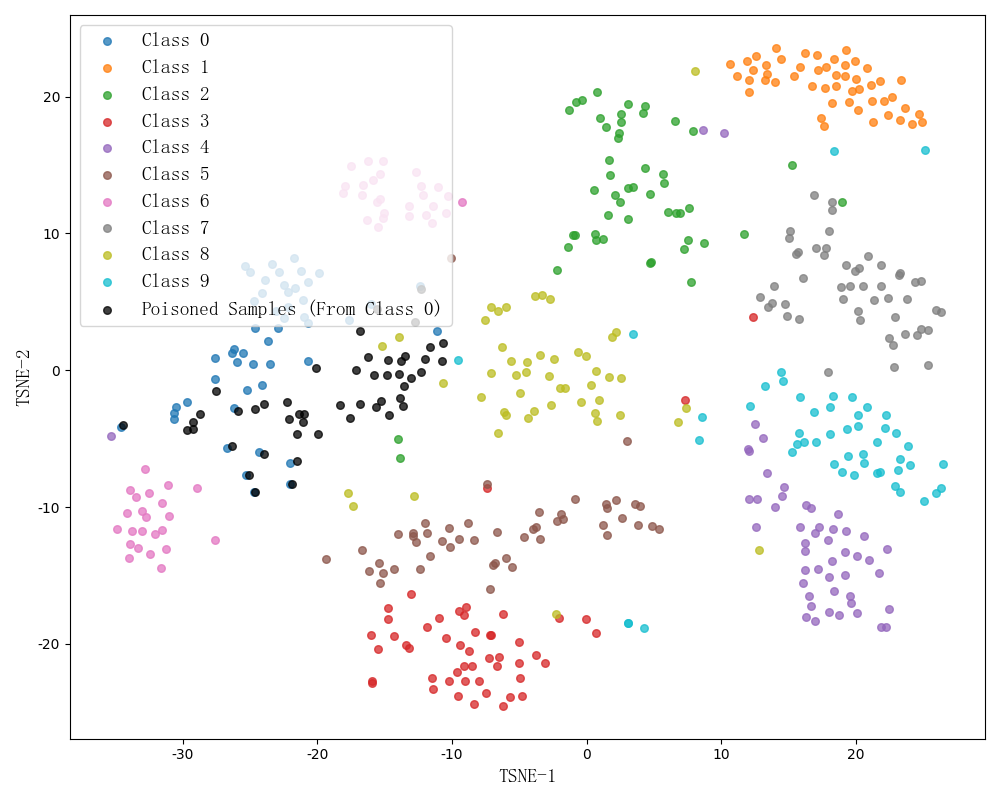}
    \caption{t-SNE visualization of the model attacked by DRL-CLBA}
    \label{fig:tsne}
\end{figure}

This section analyzes the hidden feature space behavior of DNNs under DRL-CLBA to further validate its effectiveness with t-SNE method \cite{maaten2008visualizing}. We conducted the experiment on KWS task and AISHELL3-50 dataset. The results are visualized in Figure \ref{fig:tsne}.

As shown in figure, poisoned samples (black markers, Class 0 origin) form a distinct cluster in the t-SNE projected space, significantly overlapping with yet slightly offset from the benign Class 0 cluster. This indicates that the model learns strong associations between trigger features and the target label during training. Importantly, poisoned samples remain well-separated from other classes, suggesting the attack hijacks Class 0 specifically without degrading overall discriminative capability. Meanwhile, benign samples maintain clear inter-class separability, confirming preserved generalization. However, partial Class 0 benign samples shift toward the poisoned cluster, implying increased trigger sensitivity near decision boundaries—consistent with the observed backdoor activation behavior where triggered inputs are misclassified to Class 0 regardless of true labels. These results demonstrate that DRL-CLBA effectively constructs a backdoor subspace strongly coupled with the target label while maintaining normal classification performance. The clean label strategy ensures poisoned samples remain representationally indistinguishable from benign ones, enhancing stealthiness. This feature space analysis validates DRL-CLBA's mechanism and provides insights for future defense designs.

\section{Conclusion}
In this paper, we propose DRL-CLBA, a novel clean label sample-specific backdoor attack designed for speech classification tasks. Our methodology leverages deep steganography to generate imperceptible, sample-specific triggers and utilizes a DDPG reinforcement learning framework to optimize poisoned samples through feature collision. This approach effectively eliminates the need for full gradient access to the target model and maintains the original labels of all poisoned audio samples, ensuring high stealthiness against manual inspection. Extensive experiments show that our method works. We tested it on three major speech datasets and four different AI models. In every case, the attack successfully fooled the models while they continued to perform normally on clean data. We also proved that our method can bypass common security defenses like fine-tuning and model pruning. This work highlights a serious security risk for modern voice-controlled systems.

\bibliographystyle{elsarticle-num}
\bibliography{Manuscript}

\end{document}